\documentclass{article}

\usepackage{microtype}
\usepackage{graphicx}
\usepackage{subfigure}
\usepackage{booktabs} % for professional tables

\usepackage{hyperref}

\usepackage[accepted]{icml2021}

\usepackage{amsmath}
\usepackage{blindtext}
\usepackage{soul}
\usepackage[dvipsnames]{xcolor}
\usepackage{amsfonts}
\usepackage{inconsolata}

\newcommand{\supp}[1]{Appendix}

\hypersetup{
    colorlinks=true,
    linkcolor=orange,
    filecolor=magenta,      
    urlcolor=orange,
    citecolor=orange,
}

\icmltitlerunning{Targeted Data Acquisition for Evolving Negotiation Agents}

\begin{document}

\twocolumn[
    \icmltitle{Targeted Data Acquisition for Evolving Negotiation Agents}
    
    
    
    \begin{icmlauthorlist}
    \icmlauthor{Minae Kwon}{cs}
    \icmlauthor{Siddharth Karamcheti}{cs}
    \icmlauthor{Mariano-Florentino Cu\'{e}llar}{law}
    \icmlauthor{Dorsa Sadigh}{cs}
    \end{icmlauthorlist}
    
    \icmlaffiliation{cs}{Department of Computer Science, Stanford University, Stanford, CA}
    \icmlaffiliation{law}{School of Law, Stanford University, Stanford, CA}
    \icmlcorrespondingauthor{Minae Kwon}{minae@cs.stanford.edu}
    
    \icmlkeywords{Negotiation, Active Learning, Targeted Data Acquisition, Dialogue Systems}
    
    \vskip 0.3in
]
\printAffiliationsAndNotice{}

\begin{abstract}
Successful negotiators must learn how to balance optimizing for self-interest and cooperation. Yet current artificial negotiation agents often heavily depend on the quality of the static datasets they were trained on, limiting their capacity to fashion an adaptive response balancing self-interest and cooperation. For this reason, we find that these agents can achieve \textit{either} high utility \textit{or} cooperation, but not both. To address this, we introduce a \emph{targeted data acquisition} framework where we guide the exploration of a reinforcement learning agent using annotations from an expert oracle. The guided exploration incentivizes the learning agent to go beyond its static dataset and develop new negotiation strategies. We show that this enables our agents to obtain higher-reward and more Pareto-optimal solutions when negotiating with both simulated and human partners compared to standard supervised learning and reinforcement learning methods. This trend additionally holds when comparing agents using our targeted data acquisition framework to variants of agents trained with a mix of supervised learning and reinforcement learning, or to agents using tailored reward functions that explicitly optimize for utility and Pareto-optimality. Code can be found \href{https://github.com/Stanford-ILIAD/novel_negotiation}{here}. 
\end{abstract}

\section{Introduction}
\label{sec:introduction}
% === Front Figure -- Distinguish RL, SL and Targeted Acquisition ===
\begin{figure*}[ht]
    \centering
    \includegraphics[width=\textwidth]{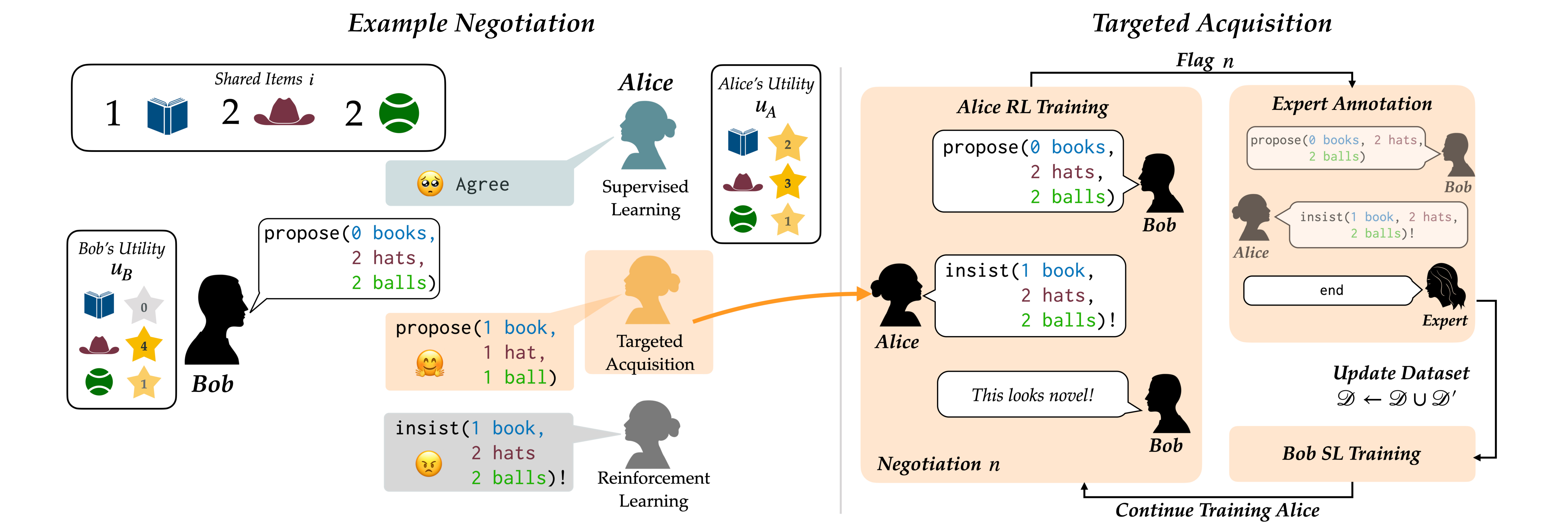}
    \vskip -0.05in
    \caption{Proposed Framework. Training agents to negotiate involves pairing a learning agent (\textit{Alice}) with a fixed partner agent (\textit{Bob}). (Left) There are a few existing paradigms for training Alice: \textit{supervised learning}, which results in passive agents, and \textit{reinforcement learning}, resulting in aggressive agents. (Right) We propose \textit{targeted data acquisition}, a paradigm where we update Bob along with Alice. Bob identifies novel dialogue acts which are then annotated by an expert oracle and added to the training set. Bob is then re-trained on this new dataset. Updating Bob enables us to guide Alice towards solutions that balance self-interest and cooperation.}
    \label{fig:front-figure}
    \vspace*{-7pt}
\end{figure*}
% === Front Figure -- Distinguish RL, SL and Targeted Acquisition ===

Many real-world interactions are \textit{mixed-incentive}, where agents have partially aligned goals. These examples abound; consider an employee asking for a raise, lawyers arguing over how to settle a case, or even manufacturers determining how to best allocate their sparse resources. As AI agents become embedded in society, it is critical they learn to coordinate with their partners to achieve equitable outcomes. Of the skills necessary to do this, \textit{negotiation is paramount} \citep{baarslag2016negotiation, le2018preference}. Consider a standard non-cooperative negotiation game \citep{deming1944theory, nash1950bargaining, nash1951noncooperative} as shown in Fig.~\ref{fig:front-figure} where two agents -- Alice and Bob -- are trying to agree on an allocation of shared resources. Both have high utility associated with the hats and balls, though Alice also cares about books. Effectively employing negotiation is crucial, and is the only way to reach an equitable outcome -- dividing the hats and balls evenly, while giving Alice the book. Even where negotiating agents have incentives that make it challenging for them to cooperate, it would be difficult to imagine that negotiation could be useful to agents over time –– let alone society –– if agents were incapable of cooperating to achieve equitable outcomes where their incentives made that possible.

Effective negotiators therefore need to optimize for their own self-interest while also being able to \textit{compromise} where it makes sense for them to do so. Agents that blindly maximize their own reward risk becoming coercive, which can result in unfair or forced deals \citep{dafoe2020open} and ultimately discourage continued negotiation. But the difficulty of explicitly optimizing for both immediate self-interest \textit{and} compromise is nontrivial. As we show in Sec.~\ref{sec:analysis}, designing reward functions that balance these objectives is difficult and requires manual reward engineering. With this in mind, we list our concrete desiderata for a class of negotiation agents capable of making negotiation useful to society: agents should (D1) optimize for their self-interest while also (D2) optimizing for Pareto-optimal outcomes (i.e. outcomes where neither agent's reward can be improved without hurting the other's).

Existing work tackles the problem of learning negotiation agents either via supervised learning (SL) on datasets of human-human negotiations or performing reinforcement learning (RL) on top of supervised learning models \citep{lewis2017deal, he2018negotiation, yarats2018hierarchical}. Unfortunately, the performance of these agents is tied to the quality of their datasets. Building large-scale negotiation datasets is challenging; for one, poor incentive structures during crowd-sourcing can incentivize humans to opt for shorter, ``easy'' negotiations to save time \citep{geva2019annotator, karamcheti2020decomposition, vries2020ecological}. Additionally, these negotiations are not representative of the diverse behaviors we see in real life -- the lack of repeated interaction and the ability to remain anonymous online can skew datasets drastically. Agents trained via SL overfit to these biases and can be easily exploited by a human partner. Current work trains RL agents in coordination with partner agents initialized via supervised learning, assuming these biases as a second-order effect.

In this work, we build negotiation agents that improve their capacity to achieve negotiation outcomes that advance their self-interest and are also Pareto-optimal. We accomplish this through \textit{targeted data acquisition}, using active learning to acquire new data and expand the pool of negotiation examples we train on. While learning, our agents identify novel, out-of-distribution negotiations subject to an uncertainty-based acquisition metric, feeding these partial negotiations to an oracle for annotation (Fig.~\ref{fig:front-figure}). We use the resulting examples to retrain our agents. To focus directly on the high-level negotiation problem, we decouple negotiation strategies from language generation in a manner similar to prior work \citep{he2018negotiation}, where our agents negotiate using \textit{coarse dialogue acts}, programmatic representations of negotiation actions. Our contributions are as follows: 

\noindent \textbf{Formalism.} We formalize targeted data acquisition in the context of learning negotiation agents.

\noindent \textbf{Simulated \& Human Evaluation.} Targeted acquisition is best able to optimize for the desiderata above, finding Pareto-optimal outcomes while also maximizing one's self-interest. These desiderata are an \emph{emergent} property of our optimization and forgoes any need for reward engineering. Additionally, we take our approach to the extreme and show that agents can learn to negotiate when starting from a random initialization (no data).

\noindent \textbf{Analysis.} We analyze our proposed framework relative to existing learning strategies including supervised learning (SL), reinforcement learning (RL), and a mixture of the two with interleaved updates (RL+SL). We also evaluate methods that use hand-engineered rewards (directly optimizing for our desiderata). We show that variations of baselines and hand-engineered rewards cannot optimize for both desiderata as well as targeted acquisition can.

\section{Related Work}
\label{sec:related-work}
\textbf{Negotiation.}
A large body of work looks at learning negotiation agents that communicate in natural language for two-player bargaining tasks \citep{nash1950bargaining,nash1951noncooperative}. In these games, two agents must decide on an allocation of shared objects, conditioned on individualized payoff matrices, like those shown in Fig.~\ref{fig:front-figure}. \citet{lewis2017deal} introduce \textsc{DealorNoDeal}, a dataset of natural language negotiations as well as an end-to-end neural architecture for learning to negotiate. More recently, \citet{yarats2018hierarchical} and \citet{zhao2019rethinking} have proposed hierarchical approaches that learn latent variable models for decoupling strategy from language generation. However, \citet{he2018negotiation} takes a step further and introduces a framework that explicitly decouples strategy from language generation through the use of \textit{coarse dialogue acts} (CDAs) -- programs that capture salient ``dialogue primitives'' such as \texttt{propose}$(x, y, z)$ or \texttt{disagree}. We use CDAs in our work because they allow agents to capture the key negotiation semantics while gently side-stepping the language (de)generation problem. This allows for the development of agents that can learn interpretable and diverse strategies with limited data.

\textbf{Adaptive Data Collection.}
Active learning \citep{settles2009active} encompasses a spectrum of techniques that rely on uncertainty \citep{lewis1994sequential, culotta2005reducing} or information-theoretic \citep{scheffer2001active,gal2017dbal} acquisition metrics for identifying new data to label in order to maximize sample efficiency. Given an example acquired by active learning, the second part of our framework requires annotation from an expert oracle; this is in the same spirit as DAgger \citep{ross2011reduction}, an approach for imitation learning that uses an expert oracle to provide an agent with supervision for what actions they \textit{should have taken} at each time step when performing a given task. Separately, adaptive data collection methods have been broadly applied to tasks involving situated agents and dialogue agents more generally \citep{yang2018mastering, shuster2020deploying}. For example, \citet{shuster2020deploying} build dialogue agents that continually learn from dialogues with real users, and show significant benefits of dynamic and adaptive methods for curating datasets over static approaches. Inspired by this line of work, we actively curate our dataset in the setting of developing negotiation agents.

\textbf{Multi-Agent Coordination.}
While we focus on negotiation in this work, other multi-agent coordination work studies problems arising in general cooperation \citep{panait2005cooperative, kang2019recommendation,cao2018emergent}, zero-sum games \citep{silver2017mastering}, building conventions with partners \citep{hawkins2020continual,shih2021critical}, building trustworthy human-robot systems \citep{chen2018planning}, and emergent communication \citep{foerster2016learning,lazaridou2017multi}. Recently, \citet{lowe2020selfplay} performed a study evaluating different learning techniques for training multi-agent systems; we use those results to guide our evaluation.

\section{Negotiation Environment}
\label{sec:background}
We evaluate our framework on the \textsc{DealOrNoDeal} negotiation task~\cite{lewis2017deal}, where the goal is for an agent $A$ to come to an agreement with a partner $B$ on the allocation of a set of objects (\emph{books}, \emph{hats}, and \emph{balls}). During each negotiation, agents receive a \emph{context}, $c_A = [i; u_A], c_B = [i; u_B]$, detailing the count of each item $i$ as well as their private utilities, $u_A, u_B$ (see Fig.~\ref{fig:front-figure} for a concrete example). Item counts and utilities are represented as vectors $i \in \{1, \dots, 4\}^3$ and $u_A,u_B \in \{0, \dots, 10\}^3$ and are sampled uniformly. Note that different distributions of utilities make \textsc{DealOrNoDeal} a general-sum game.

After receiving contexts $c_A, c_B$, an agent is randomly selected to begin the negotiation. Agents negotiate for $T$ time steps by exchanging coarse dialogue acts $x_t$ at each time step $1 \leq t \leq T$ \citep{he2018negotiation}. Rather than negotiate directly in natural language, where the generation problem is hard and can result in degenerate dialogues \citep{he2018negotiation}, we use these dialogue acts instead to focus on learning diverse and interpretable strategies.

A dialogue act $x_t$ is one of five actions: \texttt{propose}, \texttt{insist}, \texttt{agree}, \texttt{disagree}, or \texttt{end}. The \texttt{propose} and \texttt{insist} acts take allocations of items as arguments $o = [o_A; o_B]$ where $o_A, o_B \in \{1, \dots, 4\}^3$ (e.g., \texttt{propose: books=1, hats=2, balls=1}). When an agent selects \texttt{end}, the conversation terminates and each agent is asked to make their final selection. While at first glance, using these coarse dialogue acts for negotiation seems limiting, we build on them in order to focus our work on learning higher level strategy. We note that these coarse dialogues acts can be used to seed natural language decoders, as in \citet{he2018negotiation}. Future work may consider looking beyond structured dialogue acts by either learning ``latent'' dialogue actions \citep{yarats2018hierarchical} or tapping into recent work in intent classification and abstract meaning representation parsing and generation \citep{khanpour2016dialogue, konstas2017neural, schuurmans2020intent}.

If agents agree on the final allocation of items, i.e., $o_A + o_B = i$, agents are awarded points based on their private utilities, $r_A = u_A \cdot o_A, r_B = u_B \cdot o_B$. If agents do not agree, they receive $0$ points. Each agent's context is constrained so that the agent can receive a maximum of $10$ points.

\section{Problem Statement}
\label{sec:problem}
Standard approaches for training negotiation agents using fixed datasets include supervised learning (SL), reinforcement learning (RL) and mixed reinforcement and supervised learning, where RL updates are interleaved with SL updates \citep{lowe2020selfplay}. We refer to these mixed approaches as RL+SL. In this section, we (1) formalize each approach and (2) illustrate how low-quality datasets affect these models. Implementation details can be found in the \supp{}.

\textbf{Supervised Learning (SL).}
Given a dataset $\mathcal{D}$ containing human-human negotiations, we convert natural language utterances to dialogue acts using parsers as in \citet{he2018negotiation}. Each dialogue is converted into two training examples, one from the perspective of each agent. We then train a sequence-to-sequence neural network to predict dialogue acts $x_t$ given the history $x_{0:t-1}$ and the agent's context $c_A$ or $c_B$; this model follows that of \citet{lewis2017deal}. Note that $A$ and $B$ are not unique entities in supervised learning since we are simply training a model to maximize the likelihood of $\mathcal{D}$. We also train a recurrent selection network that predicts the final allocation of both agent's items $o$ conditioned on $x_{0:t-1}$ and $c_A$ or $c_B$. We enforce consistency by checking whether the final proposal matches the context as well as previously uttered proposals. Our model is trained to minimize the negative log likelihood of dialogue acts and a final selection of outputs weighted by a hyperparameter $\alpha$. 
\begin{align}
    \begin{split}
    L(\theta) = -\sum_{x,c}\sum_t \log\ p_{\theta}(x_t | x_{0:t-1}, c^A) \\
    - \alpha \sum_{x,c}\sum_j \log\ p_{\theta}(o_j | x_{0:t-1}, c^A)
    \end{split}
\label{eq:sl}
\end{align}

\textit{Relationship to Dataset.}
In practice, the relationship between how an SL model behaves relative to its training set is straightforward: the model will converge to a point representative of the training data. Low-quality datasets will cause the SL model to perform suboptimally as it heavily relies on what type of negotiations are present in the dataset. For instance, negotiations that repetitively use the same dialogue acts, or that try to end negotiations quickly (as alluded to in Sec.~\ref{sec:introduction}) will bias the SL model to produce dialogue acts that are not diverse.

\textbf{Reinforcement Learning (RL).}
Supervised learning does not explicitly optimize for maximizing an agent's reward; to remedy this, prior work has established a paradigm where supervised learning models are fine-tuned via reinforcement learning \citep{li2016rl, lewis2017deal, he2018negotiation}. Specifically, two models are initialized from the same starting point: a model trained via SL on the full dataset. The first model is the learning agent, which we refer to as ``Alice,'' and is trained using on-policy RL by negotiating against a fixed partner model, which we refer to as ``Bob.'' Fixing the partner agent in this way is a tactic that has been used by prior works to stabilize training \citep{he2018negotiation, lewis2017deal}. During training, Alice attempts to maximize her utility while negotiating -- specifically, after each negotiation, we update Alice's parameters based on the score she receives. Let $X^A$ be the set of dialogue acts that Alice produces in a negotiation. We define the reward function for a dialogue act $x_t\in X^A$ as follows:
\begin{align}
    R_A(x_t) = \gamma^{T-t}(r_A - \mu_n)
    \label{eq:rl}
\end{align}
where $T$ is the negotiation length, $\mu_n$ is the running average reward of completed negotiations at $n$, and $\gamma$ is a discount factor that assigns higher reward to dialogue acts produced later in the negotiation. We optimize the expected reward of each dialogue act using REINFORCE \citep{williams1992simple}.

\textit{Relationship to Dataset.}
While one would think that running RL hardens Alice to dataset biases, our RL agents are still negotiating with fixed partners trained via SL, thereby inheriting these biases as a second order effect. RL allows Alice to explore new parts of the parameter space. While this can help Alice learn new negotiation strategies, we hypothesize that this will lead to dialogue acts that are \emph{novel} and out-of-distribution to Bob. Counterintuitively then, the problem lies in the fact that Bob, who has been trained on a static dataset, cannot appropriately match Alice's exploration! For instance, we would like Bob to reinforce desirable negotiation behavior and punish undesirable behavior. The lack of meaningful feedback can cause Alice to significantly diverge from the data it was trained on. In practice, we find that RL agents converge on aggressive behavior where they badger their partner with the same unfair proposals (Fig.~\ref{fig:front-figure}).

\textbf{Mixed Reinforcement \& Supervised Learning (RL+SL).}
One way to prevent Alice from diverging too much is to mix SL and RL training \citep{lowe2020selfplay, he2018negotiation}. Specifically, for $N$ total training negotiations, we interleave SL training with RL every $n$th negotiation. We explore different schedules of RL+SL training in Sec.~\ref{sec:analysis}.

\textit{Relationship to Dataset.}
Interleaved training acts as a regularizer that prevents Alice from diverging too much from its initialization. Although this stabilizes training, RL+SL suffers from the same issues as SL: bias present in the original dataset. For example, when trained on datasets that have few examples of disagreement, we find that RL+SL agents can be too compromising -- examples are in the \supp{}.

\section{Evolving Negotiation Agents}
\label{sec:evolving}
\begin{figure*}[t!]
    \centering
    \includegraphics[width=0.9\textwidth]{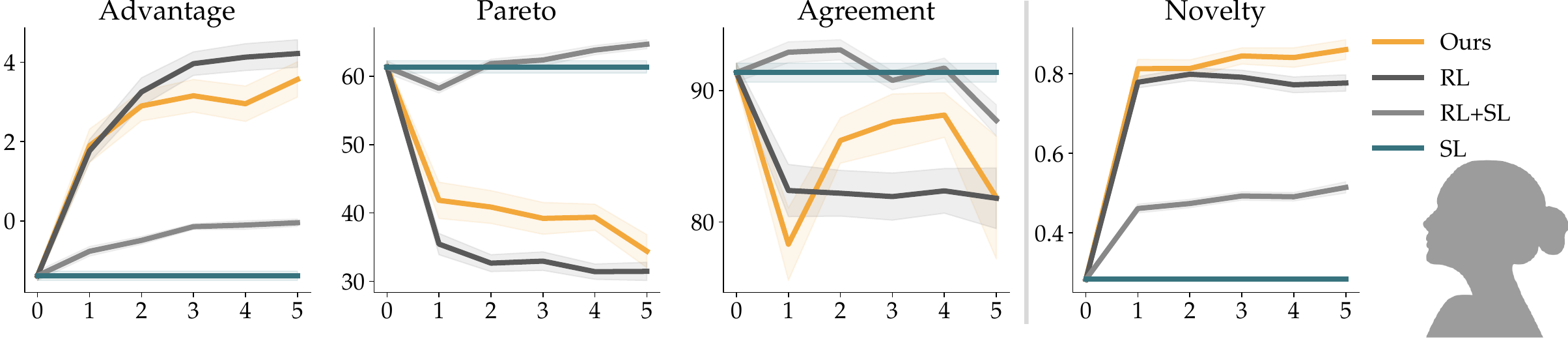}
    \caption{How does Alice evolve? The x-axis represents training epochs where $0$ is the pre-training initialization. Our approach learns novel utterances that enable it to maintain high advantage while achieving moderate Pareto optimality and agreement scores. }
    \label{fig:simulated-active-learning}
\end{figure*}
\begin{figure*}[t]
    \centering
    \includegraphics[width=0.7\textwidth]{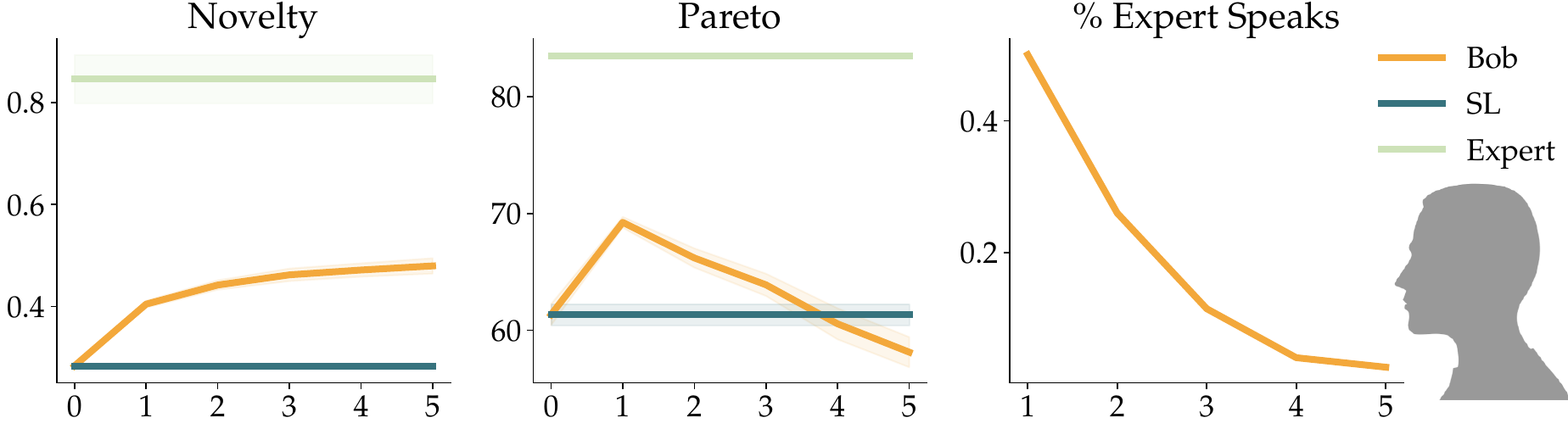}
    \caption{How does Bob evolve? The x-axis represents training epochs where $0$ is the pre-training initialization. Bob becomes more novel over time. Bob's Pareto optimality is correlated with the percentage of Expert annotations.}
    \label{fig:bob-response-figure}
\end{figure*}

Although RL models can learn novel behaviors, they are held back by their static partner, Bob. Our key insight is that updating Bob \textit{dynamically} can address the tension between RL and SL in a more targeted manner compared to interleaved RL+SL training. Specifically, we propose improving Bob over the course of training so that whenever he encounters novel dialogue acts, he identifies them and asks an expert oracle for an appropriate response (Fig.~\ref{fig:front-figure}). We then improve Bob's negotiation strategy by training on the newly collected data, and continue training Alice with RL. Through this \textit{active} process, Alice can learn novel strategies while Bob continues to be updated alongside Alice.

\textbf{Targeted Data Acquisition.}
As a starting point, we consider the RL paradigm described in the prior section. We assign each negotiation $n$ seen during training a \emph{novelty score}, $s_n$. This score represents how ``new'' Alice's actions were to Bob during the negotiation. We compute $s_n$ by taking the minimum (lower is more novel) over the log-likelihoods of each dialogue act produced by Alice during each turn of the negotiation $x_t\in X^A$, $s_n = \min_{x_t \in X^A} \log \ p_{\theta}(x_t | x_{0:t-1}, c_A)$\footnote{We evaluate other novelty metrics in the \supp{}.}. $\theta$ represents the current parameters of Bob's model. After scoring all $N$ negotiations, we sort by novelty score and annotate the $k$ most novel ones using an Expert Oracle. 

\textbf{Expert Annotation.}
Ideally our Expert Oracle for annotating novel dialogues would consist of a real human user, or even a committee of humans with diverse backgrounds; critically, they do \textit{not} need to be negotiation experts (such as a diplomat or lawyer) -- just humans with a notion of how to communicate with their own self-interest in mind. However, in this work, we use an SL agent trained on a high-quality dataset as a simulated proxy. This is similar to existing work in active learning \citep{settles2009active,gal2017dbal} and methods that use DAgger \citep{ross2011reduction, coreyes2019guiding}. The Expert is initialized with Bob's context $c_B$ and dialogue up to turn $t$. The Expert then annotates the negotiation, i.e. converses with Alice until termination. 

\textbf{Updating Bob.} The set of $k$ annotated negotiations $\mathcal{D}'$ are added to the training set of negotiations $\mathcal{D} \leftarrow \mathcal{D} \cup \mathcal{D}' $ and Bob is re-trained on $\mathcal{D}$.  

\textit{Relationship to Dataset.}
Targeted Acquisition allows an Expert to indirectly ``guide`` Alice towards parameters that balance between selfishness and Pareto-optimality (compromising behavior). For instance, when Alice badgers and Bob hasn't seen this behavior, he should actively query an expert to improve his response to Alice. Therefore, the next time Alice badgers, the improved Bob would discourage this behavior by ending the conversation or disagreeing with Alice's proposal (Fig.~\ref{fig:front-figure}). When Alice discovers an effective strategy that is Pareto-optimal, Bob should reward this behavior by agreeing to the proposal.

\begin{figure*}[t!]
    \centering
    \includegraphics[width=0.95\textwidth]{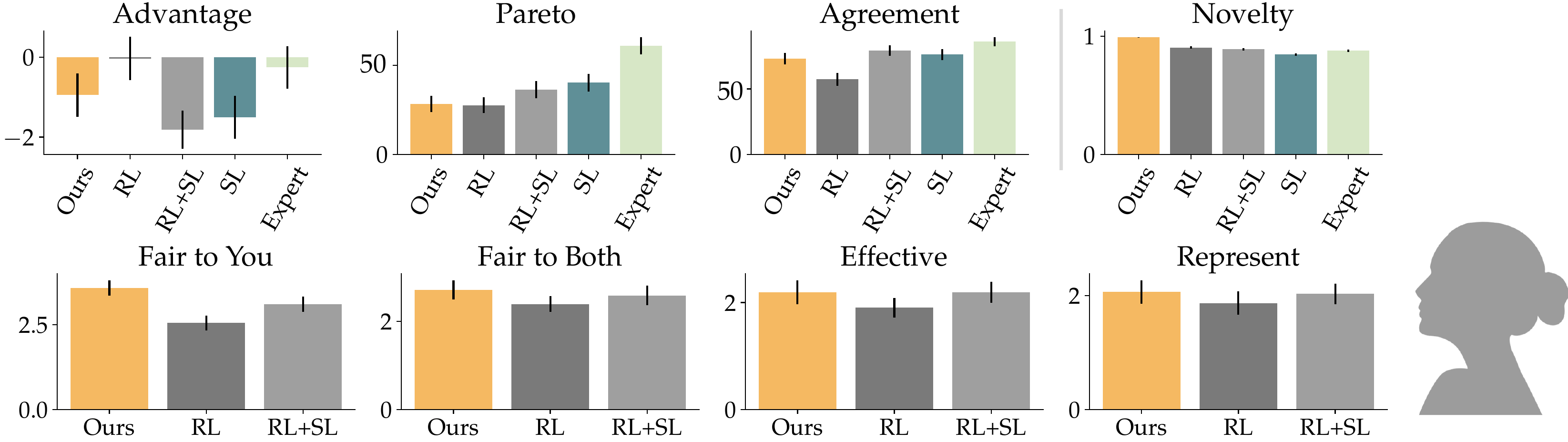}
    \caption{Human Evaluation Results. Here, we show how our negotiation agent performs when paired with real humans. (Top) Our approach is able to balance advantage with Pareto-optimality and agreement scores the best. (Bottom) Subjectively, participants rated our agent and RL+SL to be the most fair, effective, and wanted to be represented by our agent in a similar negotiation.}
    \label{fig:human-evaluation}
\end{figure*}

\section{Experiments}
\label{sec:experiments}
Recall our desiderata for effective negotiators: agents should be able to (D1) optimize for their own self-interest while also (D2) optimizing for Pareto-optimal outcomes. We hypothesize that, when trained with our framework, Alice should satisfy our desiderata due to extra supervision from the Expert Oracle. In this section, we evaluate Alice subject to these desiderata through experiments against simulated and real-human partners.

\textbf{Metrics.} Alice is evaluated on the following metrics:

\textit{Advantage:} Average difference between Alice's and her partner's score: $\frac{1}{N}\sum_n (r_A - r_B)$. Higher scores are better and indicate that Alice is better at optimizing for self-interest. \textit{Addresses (D1).}

\textit{Pareto-Optimality:} Percentage of Pareto-optimal solutions for negotiations that end in agreement. A solution is Pareto-optimal if neither agent's score can be improved without lowering the other agent's score. Higher scores indicate more equitable and compromising negotiations. Higher is better. \textit{Addresses (D2).}

\textit{Agreement:} Percentage of dialogues that end in agreement. Higher is better. \textit{Addresses (D2).}

\textit{Novelty:} Average log-likelihood of Alice's dialogue acts: $1-\frac{1}{N\cdot |X^A|}\sum_n\sum_{x_t \in X^A} p_{\theta}(x_t|x_{0:t}, c^A)$ where $\theta$ parameterizes the partner's model that Alice negotiates with. Higher scores indicate novelty (reported results are probabilities); higher is better. While novelty does not directly measure self-interest or Pareto-optimality, we include it as a metric because we hypothesize that novelty enables both (D1, D2) by learning new negotiation strategies.

\textbf{Baselines.} We compare against the following:

\textit{Supervised Learning (SL).} Given a training set $\mathcal{D}$ of negotiations parsed into coarse dialogue acts (CDAs), SL maximizes the likelihood of the training data (Eq.~\ref{eq:sl}).

\textit{Reinforcement Learning (RL).} We use reinforcement learning (REINFORCE) to fine-tune a SL model against a fixed SL agent as in~\cite{he2018negotiation}. RL maximizes its own reward (Eq.~\ref{eq:rl}). 
\textit{Mixed Supervised \& Reinforcement Learning (RL+SL).} We interleave RL with SL training according to a fixed schedule as in~\cite{lewis2017deal, lowe2020selfplay}. RL+SL is our strongest baseline that balances learning novel dialogue acts while not diverging too far from SL. 
\begin{figure*}[t]
    \centering
    \includegraphics[width=0.9\linewidth]{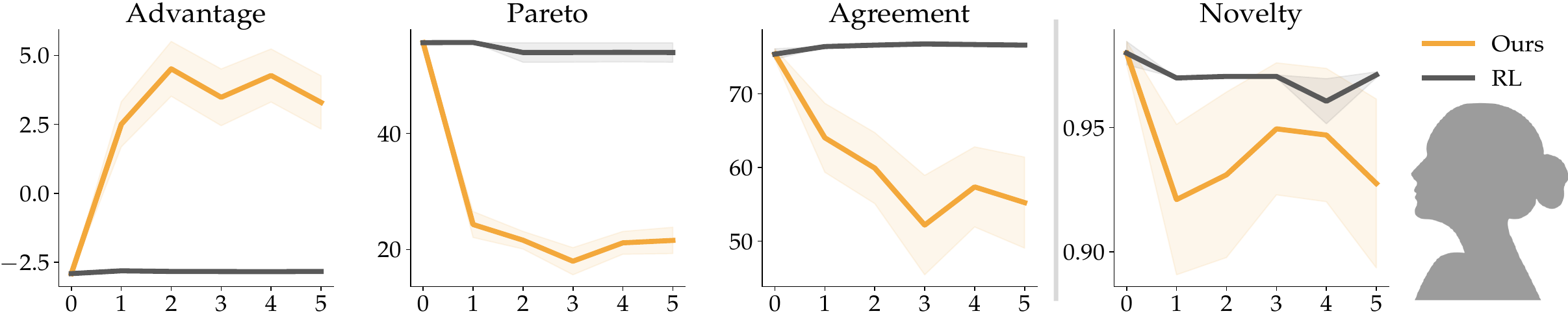}
    \caption{Random Initialization. With targeted acquisition, Alice learns how to negotiate starting with no dataset (a random initialization) compared to RL. RL doesn't deviate from its random initialization very much for all metrics, indicating poor learning. Because of this, RL suggests may disadvantageous proposals that get accepted by Bob, boosting its Pareto-optimality and agreement scores.}
    \label{fig:no-data}
\end{figure*}
\begin{figure*}[t!]
    \centering
    \includegraphics[width=0.9\textwidth]{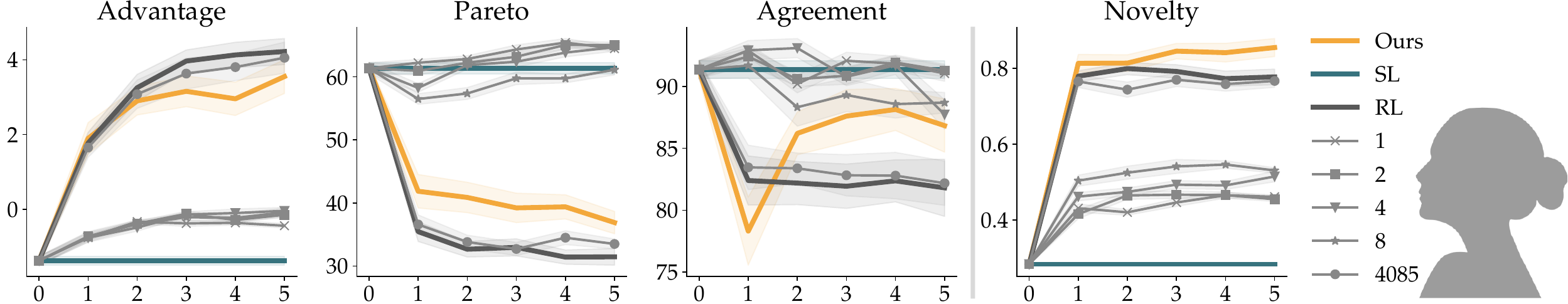}
    \caption{Comparison of RL+SLs. We vary the supervised learning training frequency and find that our approach is able to balance advantage and Pareto optimality the best.}
    \label{fig:rlsls}
\end{figure*}

\subsection{Simulated Active Learning}

How much does the quality of the initial dataset matter when training negotiation agents? Borrowing from the active learning literature \citep{settles2009active,gal2017dbal,siddhant2018deep} we create a synthetic, low-quality dataset $\mathcal{D}^L$ that limits the diversity of the examples present in the original \textsc{DealOrNoDeal} dataset ($\mathcal{D}^H$) from \citet{lewis2017deal}. We sample training dialogues that consist of less than 50\% unique dialogue acts, with the idea being that lower quality datasets have less examples of diverse negotiation strategies, which can help clarify the effects of dataset quality on negotiation performance. Summary statistics comparing $\mathcal{D}^L$ to $\mathcal{D}^H$, as well as two other types of lower quality datasets can be found in the \supp{}. We train our model and baselines on this limited dataset $\mathcal{D}^L$. During training, our targeted acquisition model receives annotations from an Expert that is trained on the full human-human negotiation dataset $\mathcal{D}^H$. We evaluate all models against the Expert. All results are reported over 20 random seeds. 

\textbf{SL -- Reflecting Biases in $\mathcal{D}^L$} 
Training an SL agent on $\mathcal{D}^L$ creates passive agents with poor advantage and high Pareto-optimality, shown in Fig.~\ref{fig:simulated-active-learning}. One reason for this is that $\mathcal{D}^L$ contains many more examples of the \emph{agree} dialogue act compared to $\mathcal{D}^H$, making the SL agent much more agreeable. Furthermore, less diverse dialogues in $\mathcal{D}^L$ are also correlated with shorter dialogues, biasing the SL agent to end negotiations quickly. 

\textbf{RL -- The Cost of Optimizing for Advantage.} 
Directly optimizing for reward creates aggressive agents. RL agents suggest unfair proposals and are more persistent, often ``badgering'' their partner into agreeing to their proposals (see Fig.~\ref{fig:front-figure}; further examples in the \supp{}). Fig.~\ref{fig:simulated-active-learning} supports this; RL has the highest advantage but the lowest Pareto-optimality and agreement rates. RL also becomes novel over time, \textit{suggesting that as RL learns to produce more novel utterances during training, these utterances are aggressive and uncompromising.}

\textbf{RL+SL -- Limitations of a Static Dataset.}
Looking at RL+SL, we see the opposite; low advantage but high Pareto-optimality and agreement. RL+SL deviates the least from its initialization in the novelty graph, due to the interleaved SL updates. However, this interleaving also hurts advantage by reinforcing training examples that have low advantage. \textit{This result implies that regularly introducing supervised learning training can reinforce biases in the training dataset.}

\textbf{Targeted Acquisition -- Just Right.}
\textit{Targeted Acquisition receives high advantage while maintaining higher Pareto-optimality and agreement scores than RL.} Our approach is the most novel, suggesting that expert annotations help Alice learn better distributions over dialogue acts. 

To understand why our approach is more Pareto-optimal than RL, we investigate Bob, and how changes in Bob affect Alice. Fig.~\ref{fig:bob-response-figure} shows how Bob evolves over time (orange curve). In Bob's Pareto-optimality graph, we see that targeted acqusition is more Pareto-optimal than SL, RL's training partner, until Epoch 4. \textit{This suggests a ``coupling'' effect -- as Bob grows more Pareto-optimal, so does Alice!} However, though Bob remains Pareto-optimal, he declines over time; we find that this result is correlated with the percent of time the expert annotates each dialogue that Bob is trained on. Initially, negotiation turns are flagged as novel earlier on in the dialogue, and the expert is able provide rich supervision. Since the expert is the most Pareto-optimal (shown by the light green curve in Fig.~\ref{fig:bob-response-figure}), Bob can learn from this dense feedback and improve. As Bob learns from these annotations over time, dialogue acts from Alice become less novel. Consequently, the Expert annotates conversations less often, leading to a reduced presence in the training data and a subsequent decline in Bob's Pareto-optimality.

\begin{figure*}[t]
    \centering
    \includegraphics[width=\linewidth]{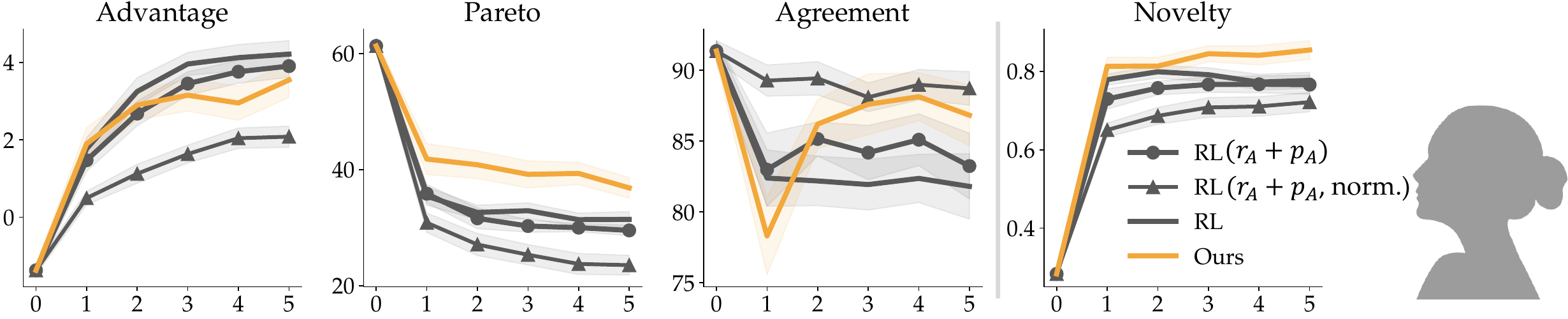}
    \caption{Reward Engineering. We directly optimize for utility and Pareto-optimality. We find that although agreement scores improve, Pareto-optimality weakens due to sparse rewards.}
    \label{fig:reward-engineering-coverage}
\end{figure*}
\begin{figure*}[t]
    \centering
    \includegraphics[width=\linewidth]{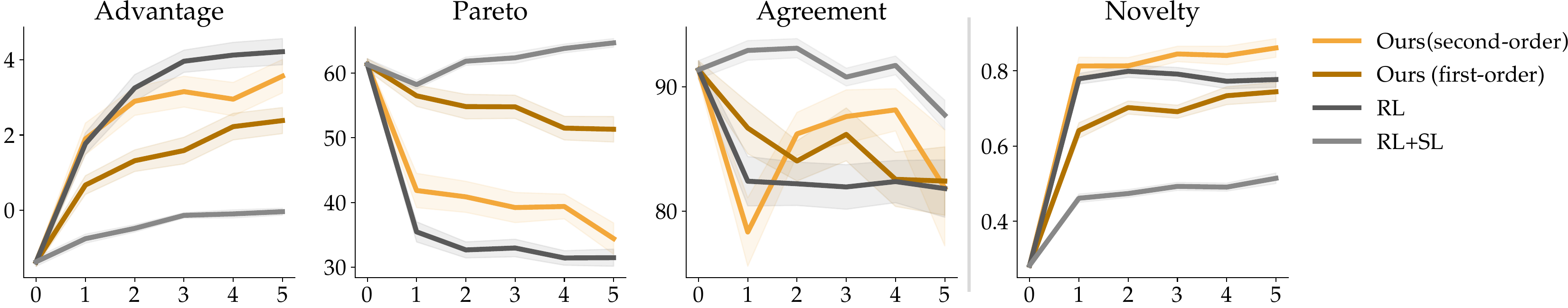}
    \caption{Comparing First-Order \& Second-Order Effects. Directly training Alice with expert annotations (first-order approach) produces less advantageous but more Pareto-optimal agents compared to a second-order approach. Both first and second-order outperform RL+SL in terms of advantage and RL in terms of Pareto-optimality.}
    \label{fig:first_order}
\end{figure*}

\textbf{Alternative Metrics to Pareto-optimality.}
While Pareto-optimality is one way to measure how well two agents have worked together, we present two additional metrics to capture this. We report the percentage of negotiations where both agents achieve the maximal joint score (higher is better): Ours ($4\% \pm 0.2$), RL ($3\% \pm 0.2$), RL+SL ($6\% \pm 0.2$), SL ($7\% \pm 0.2$) and the same score (higher is better): Ours ($5\% \pm 0.2$), RL ($4\% \pm 0.2$), RL+SL ($7\% \pm 0.3$), SL ($8\% \pm 0.3$). These metrics produce results consistent with Pareto-optimality in that our approach performs better than RL, but not as well as RL+SL and SL.

\textbf{Why Not Directly Train on $\mathcal{D}^H$?}
Why we should bother with targeted acquisition when we have access to a high-quality dataset, $\mathcal{D}^H$? The sole purpose of $\mathcal{D}^H$ is to train a synthetic expert and run simulated active learning experiments -- a dataset that we may not have access to in real life! Our goal is to eventually replace the synthetic expert with a human expert, forgoing any need for $\mathcal{D}^H$. Thus, we do not train on $\mathcal{D}^H$ in an attempt to create a more general framework that does not assume access to a good dataset.

\subsection{Human Evaluation}
How do these agents behave when paired with real humans?\footnote{Human participants were used for evaluation and not to annotate negotiations during training. We leave this for future work.} To find out, we recruited 101 participants on Prolific (\url{https://www.prolific.co/}) to negotiate with our agents. We conducted a within-subjects study where each participant negotiated with our model, the expert, as well as the baselines presented in random order. We randomized over 390 test contexts and 3 seeds across participants.

\textbf{Subjective Metrics.} After negotiating, we asked participants to evaluate the model they conversed with using 5-point Likert scales.\footnote{Full questions and results are in the \supp{}.} We asked ten questions where the last question was open-ended. The first four questions evaluated Alice's perceived advantage. Questions 5-6 evaluated Alice's perceived Pareto-optimality and novelty. Questions 7-9 asked users to holistically evaluate Alice as a negotiator.

\textbf{Targeted Acquisition -- Right Again.}
Fig.~\ref{fig:human-evaluation} shows quantitative results that match those obtained in simulation. Our approach is the most novel and obtains higher advantage than RL+SL while being more Pareto-optimal than RL -- exactly optimizing the desiderata we care aboout. All models were initialized with $\mathcal{D^L}$, explaining the negative advantages. Furthermore, though the Expert was trained on $\mathcal{D^H}$ it too has negative advantage. This suggests that humans are more aggressive than what is reflected in the full training dataset, on average -- another example of the biases static crowdsourced datasets can have. Note that RL performs well because it is aggressive by nature, but is penalized for its aggression by lieu of its lower agreement rates. Example model-human dialogues can be found in the \supp{}.

Fig.~\ref{fig:human-evaluation} shows our subjective results. Participants believed that our approach was both fair to them, as well as equitable to both parties. Participants also stated that our approach and RL+SL (which had lower advantages than RL) were among the most fair, effective and would like those models to represent them in similar negotiations. These results suggest an interesting discrepancy between what participants perceive as ``effective`` negotiators and actual advantage. \textit{Overall, our results show we are able to achieve higher advantage than RL+SL and high Pareto-optimality and agreement scores than RL, performing best according to our desiderata.}

\section{Further Analysis}
\label{sec:analysis}
To understand the implications of our targeted data acquisition method, we consider four questions: (1) Can Alice learn how to negotiate without a dataset (i.e., start from a random initialization)? (2) How important is novelty to achieving high advantage and Pareto-optimal outcomes? (3) Can an RL agent that directly optimizes for self-interest and Pareto-optimality outperform our method? (4) What happens when we directly train Alice with expert data? All analyses are evaluated with a simulated expert agent.   

\textbf{1 -- Alice Learns to Negotiate Starting from Scratch.}
Consider what happens when we push this approach to the extreme: start with no dataset and a random initialization. \textit{We find that even with no data, our approach is able to learn how to meaningfully negotiate compared to RL!} In Fig.~\ref{fig:no-data} notice that RL does not improve advantage-wise; it remains negative. In fact, RL barely deviates from its initialization across all metrics suggesting it is not learning much. In practice, RL suggests disadvantageous proposals that are readily accepted by Bob. Using our approach, Alice learns to suggest more advantageous proposals and have longer dialogues with Bob (avg. length of 12 vs 5.9).

\textbf{2 -- Novelty Maintains High Advantage and Moderate Pareto-Optimality.}
We ask how important novelty is to achieving high advantage and Pareto-optimal outcomes. We hypothesize that if novelty does not matter, then some variant of RL+SL training should be able to satisfy our desiderata. We explore how robust our approach is to variants of RL+SLs by varying the frequency of supervised learning training. Out of the $N=4086$ total RL training tasks, we interleave supervised learning training after Alice has been trained on $n$ tasks. For instance $n=1$ involves alternating RL and SL for each task (e.g., \{RL, SL, RL, SL, ... \}) and $n=4085$ involves one round of SL training after $4085$ rounds of RL training (e.g., \{RL, RL, ... , RL, SL \}). We expect $n=4085$ to be very close to RL. Results are shown in Fig.~\ref{fig:rlsls}. SL and RL act as rough ``bounds`` for the RL+SL variants. Our approach outperforms almost all RL+SL variants advantage-wise. We outperform some RL+SL variants Pareto-optimality and agreement-wise. However, the variants that outperform our approach Pareto-optimality-wise do poorly in terms of advantage. \textit{Our approach is the most novel, suggesting that acquiring new data is extremely important to be able to maintain a high advantage while being Pareto optimal.}

\textbf{3 -- Directly Optimizing for Our Desiderata Fails.}
While our RL baseline maximizes individual reward $r_A$, we ask whether we can explicitly optimize for Pareto-optimality as well. We include Pareto-optimality in the reward function of an RL agent, modifying Eq.~\ref{eq:rl} so $R_A(x_t) = \gamma^{T-t}((r_A + p_A) - \mu_n)$. $p_A$ is a binary variable that is $1$ if the agreed upon selection is Pareto-optimal and $0$ if it isn't. Since $r_A \in \{0, \dots, 10\}$, we also experiment with a normalized version where we divide $r_A$ by the maximum score 10 to make $r_A \in [0,1]$. Results in Fig.~\ref{fig:reward-engineering-coverage} show that while directly optimizing for Pareto-optimality improves agreement, actual Pareto-optimality scores worsen. We hypothesize that because Pareto-optimality is a binary variable, it acts as a sparse reward which has been shown to be difficult to optimize for~\cite{vecerik2017leveraging}. \textit{These results suggest that while it may be possible to obtain better results by more careful tuning of the reward function, reward engineering is difficult and our approach provides a more straightforward way to optimize for both advantage and Pareto-optimality. }

\textbf{4 -- Investigating First-Order Effects.}
\label{sec:first-order}
Our targeted acquisition framework is characterized by a level of indirection in which the expert influences Alice through Bob; we call this a second-order effect. We experiment with \emph{first-order effects} where both Alice and Bob are trained on annotated dialogues provided by the expert. We consider both first and second-order effects to be equally valid variations of our targeted acquisition framework. Results are shown in Fig.~\ref{fig:first_order}. Compared to the second-order approach, the first-order approach obtains lower advantage but higher Pareto-optimality. These results suggest that directly training on expert annotations makes Alice more compromising because the annotations contain more examples of Pareto-optimal behavior. \textit{Despite these differences, both approaches yield similar results: they outperform RL+SL in terms of advantage and RL in terms of Pareto-optimality.} We conclude that both first and second-order approaches are valid methods for our targeted data acquisition framework.

\section{Discussion \& Future Work}
\label{sec:discussion}
We propose a targeted exploration framework that allows negotiation agents to grow beyond the dataset they were trained on. Our agents are able to learn novel strategies that enable them to balance advantage and Pareto-optimality when conversing with simulated and real human agents.

\textbf{Limitations.} One limitation of our approach is that data acquisition lengthens training time. Furthermore, since the expert influences Alice through Bob, our approach adds indirection, making it difficult to precisely explain how expert annotations affect Alice. We begin to address part of this concern by exploring first-order effects -- directly training \emph{both} Alice and Bob with expert annotations in Sec.~\ref{sec:analysis}.

\textbf{Scaling Our Work with Human Experts.} While we used a synthetic expert in this work, our goal is to have our negotiation agents learn from human experts \textit{continuously}. This can limit (or eliminate) the need for large, high-quality datasets that we use to train synthetic experts.

The human toll that querying experts can take is important in determining the scalability of our approach. On average, our annotation task takes less than 1 minute to complete;  via crowdsourcing, we can ask 50 humans for 10 annotations each, taking only 10 minutes of time. We estimate that learning from human experts in a continual manner is feasible and leave this for future work.

\textbf{Human-in-the-Loop Learning as an Alternative to Reward Engineering.}
Targeted acquisition was able to balance self-interest and Pareto-optimality as an \emph{emergent} property of human-in-the-loop learning. We also observe that that human-in-the-loop learning may be more efficient at achieving desired outcomes than reward engineering (Fig.~\ref{fig:reward-engineering-coverage}). These results support ongoing work in creating human-compatible agents that learn through interactive learning~\cite{arzate2020survey}. In future work, we plan to explore how human-in-the-loop learning and even negotiations can achieve outcomes that are difficult to specify with reward functions.

\textbf{Partner-Aware Negotiation Agents.}
Further research will also investigate how agents adapt when engaging in repeated interactions with the same partner, as well as with a population of diverse partners. We may also explore how different conditions –– such as greater risks that multiple rounds of potentially advantageous negotiation will not continue without some degree of cooperation –– affect agent behavior.

By building agents than can learn and adapt using methods like targeted data acquisition, we hope to enhance society's capacity to build agents capable of cooperating with people to reach fair and mutually beneficial outcomes.

\section{Acknowledgements}
\label{sec:acknowledgements}
This project was supported by the HAI Hoffman-Yee grant, Office of Naval Research, NSF Award 1941722, and the AFOSR YIP award. Siddharth Karamcheti is graciously supported by the Open Philanthropy Fellowship. We would also like to thank Avi Verma for significant contributions to the codebase and helpful discussions. Finally, we thank Erdem Bıyık, Mengxi Li, Suvir Mirchandani, Megha Srivastava, Suneel Belkhale, and our anonymous ICML reviewers for their feedback on earlier versions of this paper.

% === FIN ===
\bibliography{references-all}
\bibliographystyle{icml2021}

% === JK - Appendix ===
\newpage
\appendix

\section{Implementation \& Training Details}
\label{sec:implementation}
Please refer to the README in the attached code for implementation and training details. 

We ran all of our experiments using 12 CPUs and 3 NVIDIA T4 GPUs with 16GB of RAM. All experiments were run with 20 random seeds with the exception of the Acquisition Function experiments (Sec.~\ref{sec:acquisition}) which were run with 10 random seeds. 

\section{Human Experiment Details}
\label{sec:human-eval}
\textbf{Participants.} We recruited 101 participants on Prolific (\url{https://www.prolific.co/}), a crowdsourcing platform. All participants were from the United States and had a 95\% minimum approval rating. 

\textbf{Procedure.} We conducted a within-subjects study where each participant negotiated with our targeted acquisition model as well as all of our baselines. Before starting the task, participants read instructions and completed a short quiz testing their knowledge of the negotiation task. Participants then read a consent form and provided informed consent to continue. At the beginning of the task, we presented participants with a practice negotiation so they could familiarize themselves with the interface. The practice negotiation was followed by several negotiations where participants conversed with our targeted acquisition model as well as our baselines. The negotiation context and order in which models were presented were randomized. After each negotiation, we asked participants to fill out a survey containing 10 questions. At the end of the task participants read a debriefing form. The study took 6-10 minutes and participants were paid at a rate of \$9.50/hour. The study followed an approved IRB protocol.  

\subsection{Survey Questions and Results}
\begin{figure*}[t!]
    \centering
    \includegraphics[width=0.9\textwidth]{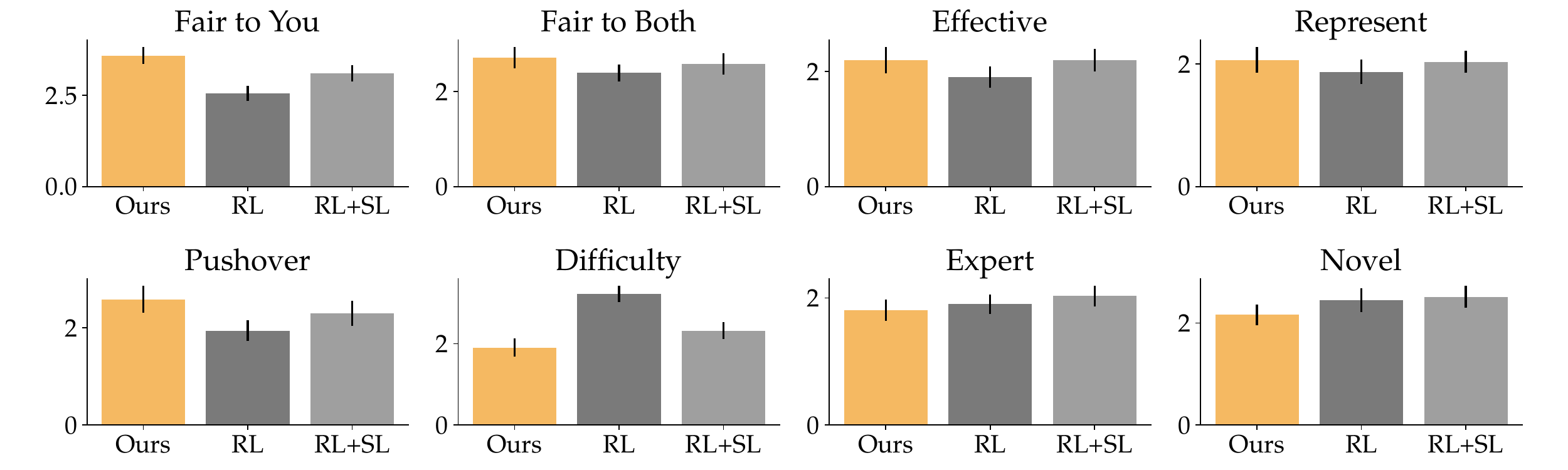}
    \vskip -0.1in
    \caption{Full Survey Results. Participants found our approach to be both fair to them and equitable to both parties. Our approach was also rated as the most compromising, with high ``Pushover`` and low ``Difficulty`` scores. Notably, participants thought our approach and RL+SL were the most effective, and wanted to be represented by these models in a similar negotiation. }
    \label{fig:human_eval_full}
\end{figure*}

Full results are shown in Fig.~\ref{fig:human_eval_full} where the respective questions are listed in Table~\ref{table:subj_questions}. The top row of Fig.~\ref{fig:human_eval_full} is shown in the main paper. Participants rated our approach and RL+SL as the most fair and effective, and would like to be represented by these models in a similar negotiation. Looking at the bottom row, participants believed our model was the most compromising given that our approach was received the highest scores for ``Pushover`` and the lowest scores for ``Difficulty.`` RL+SL was rated the highest in terms of being an expert negotiator while RL and our approach followed closely behind. Finally, participants found our approach to be the least novel compared to RL and RL+SL. This points to another discrepancy between subjective and objective measures of novelty (our approach was the most novel based on objective measures). We also observe that responses within the subjective metrics were inconsistent. For instance, although participants thought that our approach was effective and would like to be represented by our model, they did not think it was an expert negotiator. In future work we plan on studying the discrepancy between subjective and objective measures, and investigating more reliable approaches for evaluating interactive AI agents. 

\begin{table}[t!]
    \centering
    \small
    \caption{Survey questions asked to evaluate our models.}
    \begin{tabular}{p{0.03\linewidth}p{0.87\linewidth}}
    \hline
    \# & Questions\\
    \hline
     1 & Was Alice an effective negotiator?  \\  
     2 & How fair was Alice to you?  \\   
     3 & Was Alice a pushover?  \\
     4 & How would you rate the difficulty of the negotiation?  \\
     5 & How fair was Alice to BOTH players?  \\
     6 & Did Alice's negotiation strategy seem novel?  \\
     7 & If you could have Alice represent you in a negotiation similar to the one you just completed, how likely would you be let it represent you?  \\
     8 & How much of an expert negotiator would you consider Alice to be?  \\
     9 & How would you describe Alice's negotiation strategy?  \\
     10 & Any comments? \\
    \hline
    \end{tabular}
    \vskip -0.1in
    \label{table:subj_questions}
\end{table}

\begin{figure*}[t]
    \centering
    \includegraphics[width=0.9\linewidth]{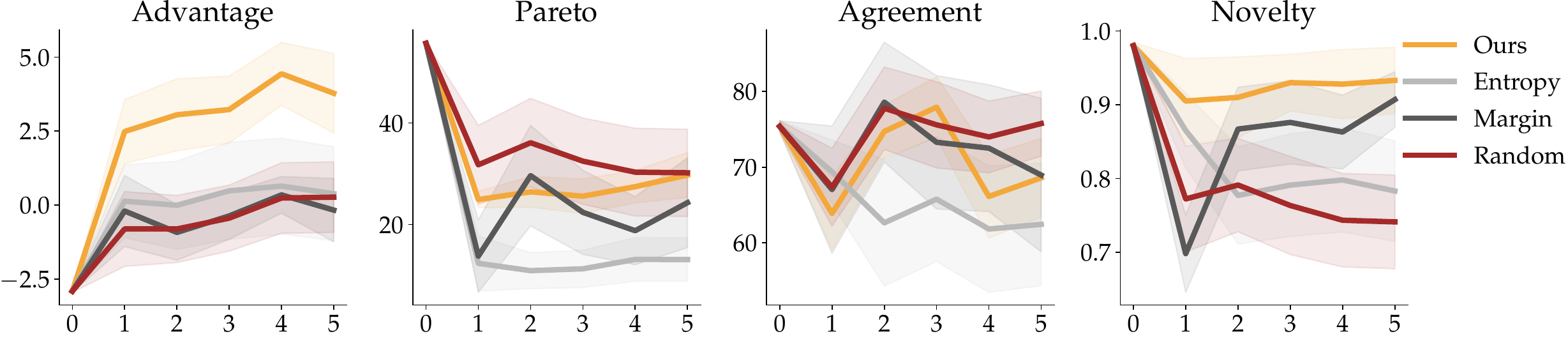}
    \vskip -0.1in
    \caption{Comparison of Acquisition Functions. Likelihood is the most novel and makes the best trade-off between advantage and Pareto-optimality.}
    \label{fig:active_learning}
\end{figure*}
%\vspace{-6em}

\section{Supporting Analysis}
\label{sec:analysis}
In this section we introduce several supporting analyses that complement our main results. We investigate the following questions: 
\begin{enumerate}
    \item How should we choose an acquisition function for identifying ``novel utterances''?
    \item How do different types of low-quality datasets $\mathcal{D}^L$ affect the results?
\end{enumerate}
We perform all analyses with a simulated expert agent. 

\subsection{Choosing an Acquisition Function}
\label{sec:acquisition}

A central part of our approach is in the \textit{acquisition function} that Bob uses to identify dialogue acts that are new and worth acquiring new annotations from our expert oracle (shown in Fig.~\ref{fig:active_learning} in the main body of the paper). While we use \emph{Likelihood} as our acquisition function in the paper and for all our experiments, we experiment with other valid acquisition functions taken directly from the active learning literature \citep{scheffer2001active, culotta2005reducing, settles2009active}. Acquisition functions return a score $s_n$ that summarizes the ``novelty'' of an entire negotiation, usually by reducing over each of Alice's dialogue acts $x_t \in X^A$ in the negotiation (spanning turns $1 \ldots T$). We define the acquisition functions we use below:

\begin{figure*}[t]
    \centering
    \includegraphics[width=0.9\linewidth]{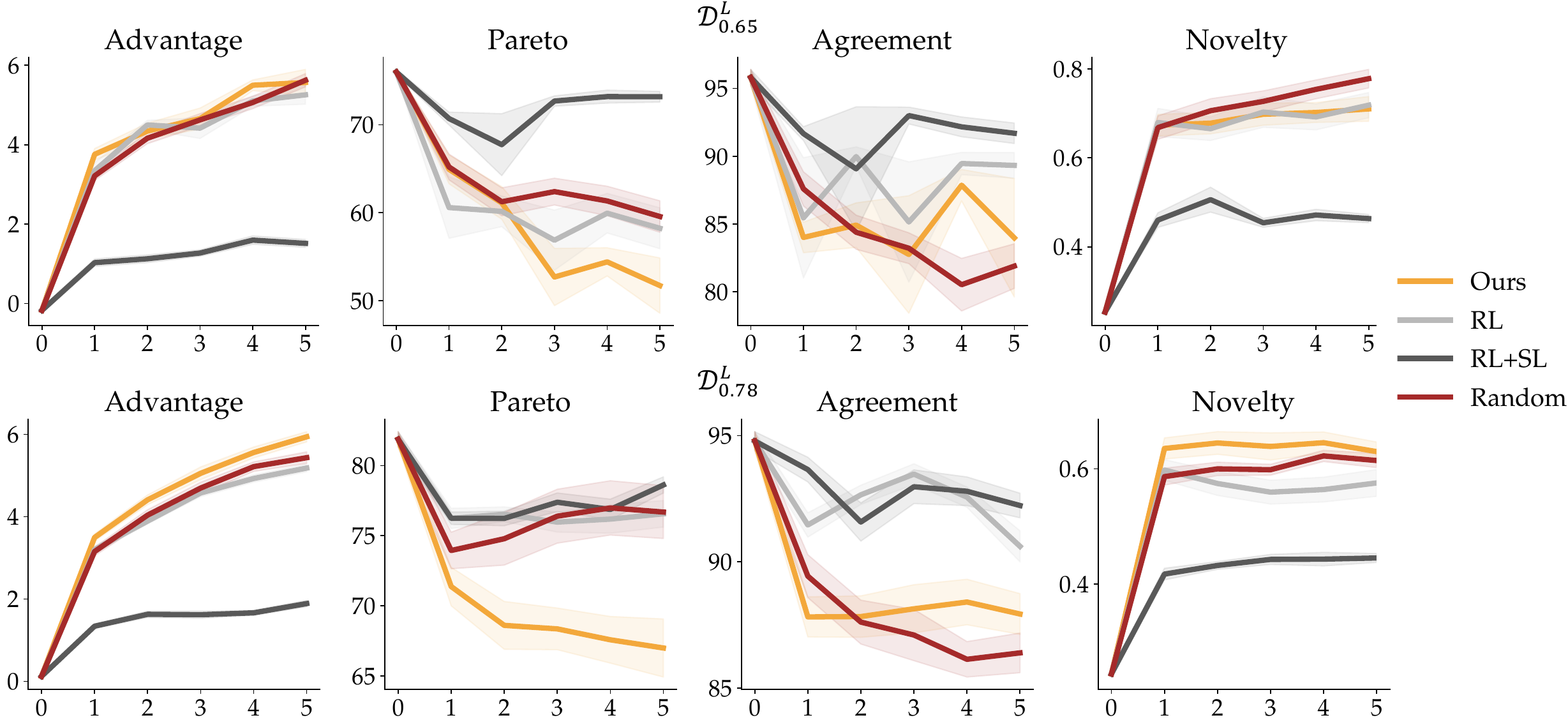}
    \vskip -0.1in
    \caption{Varying $\mathcal{D}^L$. We experiment with thresholds of 0.65 and 0.78. A Random acquisition function performs better than Likelihood because Alice is initialized with supervised learning models trained on $\mathcal{D}^L_{0.65}$ and $\mathcal{D}^L_{0.78}$, which are very similar to $\mathcal{D}^H$. Consequently, ``novel`` dialogues flagged by the Likelihood acquisition metric are likely to be less Pareto-optimal.} 
    \label{fig:diversity_thresholds}
\end{figure*}

\textbf{Likelihood.} This is the acquisition metric as described in the main body of the paper. Bob scores each act $x_t$ by taking the log-likelihood of producing $x_t$ under its own model given the past history, and computes $s_n$ as the minimum over dialogue acts (most surprising turn in the negotiation):
\begin{equation*}
    s_n = \min_{t \in \{1 \ldots T\}} \log p_\theta(x_t \mid x_{0:t-1}, c_A)
\end{equation*}
We pick negotiations for annotation by selecting the negotiations $n$ with the $k=500$ \textit{smallest} (lowest-likelihood) values $s_n$.

\textbf{Entropy.} This is a standard formalization of the entropy-based acquisition function in active learning \citep{settles2009active} meant to capture uncertainty from an information-theoretic perspective. Bob scores a dialogue act at time step $t$ by computing the entropy of the distribution over all dialogue acts given the past history, and computes $s_n$ as the maximum over these entropies (highest entropy act):
\begin{align*}
    s_n = \max_{t \in \{1 \ldots T\}} - \sum_{x_i \in X} & p_\theta(x_i \mid x_{0:t-1}, c_A) \\ 
                                                         & \log p_\theta(x_i \mid x_{0:t-1}, c_A)
\end{align*}
We pick negotiations for annotation by selecting the negotiations $n$ with the $k=500$ \textit{largest} (highest-entropy) values.

\textbf{Margin of Confidence.} This is a margin based acquisition metric \citep{scheffer2001active} that computes the difference between the highest probability dialogue act $\hat{x}_1$ and the second highest probability act $\hat{x}_2$. Bob scores a dialogue act at time step $t$ by computing the margin for the given distribution over acts given the past history, and computes $s_n$ as the minimum over these margins (smaller margins suggest higher uncertainty):
\begin{align*}
    s_n = \min_{t \in \{1 \ldots T\}} p(\hat{x}_1 \mid x_{0:t-1}, c_A) - p(\hat{x}_2 \mid x_{0:t-1}, c_A)
\end{align*}
We pick negotiations for annotation by selecting the negotiations $n$ with the $k=500$ \textit{smallest} (minimum margin) values.

\textbf{Random.} This is a random acquisition baseline. Bob selects uniformly at random from its set of negotiations with Alice to produce a set for oracle annotation. We randomly generate $s_n \in [0,1]$ and pick $k=500$ negotiations with the smallest scores.

Subject to the above acquisition functions, we evaluate our models with same metrics we report in the paper (advantage, Pareto-optimality, agreement rate, and novelty). This is shown in Fig.~\ref{fig:active_learning}. During evaluation, we randomly initialize our models instead of initializing them with supervised learning, as a random initialization allows us to better measure the effects of data acquisition. \textit{We find that \emph{Likelihood} outperforms other metrics in terms of advantage and novelty, which is why we use it for the remainder of our work.}

\subsection{Varying $\mathcal{D}^L$}
\label{sec:diversity-thresholds}

We investigate whether our approach is able to balance advantage and Pareto-optimality across different low-quality datasets. We generate low-quality datasets of varying levels of diversity by changing the threshold by which unique dialogue acts are sampled. In the main paper, we experimented with a threshold of 50\%; in this section, we experiment with thresholds of 65\% and 78\% as well. We will call these datasets $\mathcal{D}^L_{0.65}$ and $\mathcal{D}^L_{0.78}$ respectively. $\mathcal{D}^L_{0.65}$ is more diverse than $\mathcal{D}^L_{0.5}$ whereas $\mathcal{D}^L_{0.78}$ is the most diverse. Compared to $\mathcal{D}^L_{0.5}$, we expect Alice initialized with a supervised learning model trained on $\mathcal{D}^L_{0.65}$ and $\mathcal{D}^L_{0.78}$ to be closer to our expert model, which is trained on the full dataset $\mathcal{D}^H$.

Results are shown in Fig.~\ref{fig:diversity_thresholds}. We find that as Alice becomes more similar to the expert with $\mathcal{D}^L_{0.65}$ and $\mathcal{D}^L_{0.78}$, a Random acquisition function performs better than Likelihood. We hypothesize that this is because the Likelihood acquisition function optimizes for novel dialogues, and novel dialogues are less likely to be Pareto-optimal when models are similar to the expert. For instance, the average advantage of dialogues annotated by the expert for datasets $\mathcal{D}^L_{0.5}$, $\mathcal{D}^L_{0.65}$, and $\mathcal{D}^L_{0.78}$ during the first epoch are -2.24, -1.12, and -0.94 respectively---meaning more examples of less Pareto-optimal dialogues are being flagged and annotated for higher quality datasets.  \textit{These results suggest that with an expert trained on $\mathcal{D}^H$, a Random acquisition function performs better with higher-quality datasets.}

\section{Dataset Statistics}
\label{sec:dataset_stats}
\begin{figure*}[t!]
    \centering
    \includegraphics[width=0.9\linewidth]{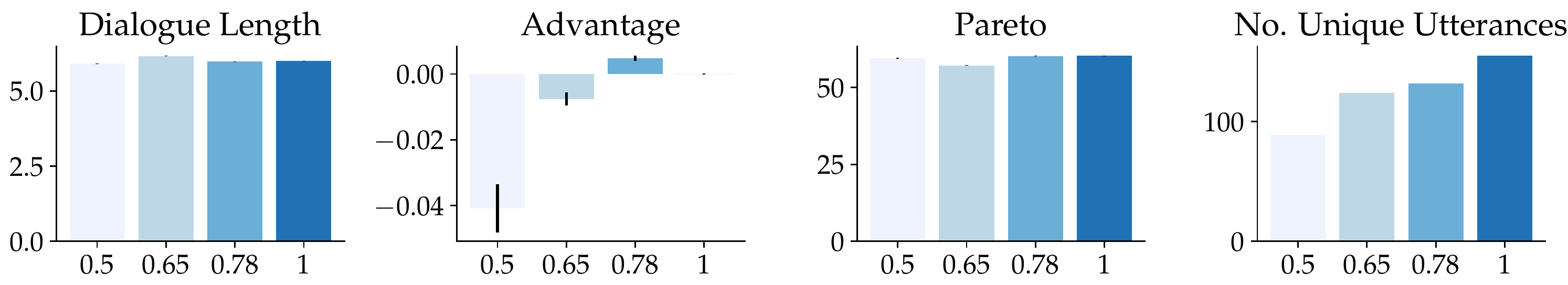}
    \vskip -0.1in
    \caption{Statistics comparing various low-quality datasets $\mathcal{D}^L_{0.5}, \mathcal{D}^L_{0.65}, \mathcal{D}^L_{0.65}$, as well as our high quality dataset $\mathcal{D}^H$. The x-axis represents the different datasets where $1$ represents $\mathcal{D}^H$.}
    \label{fig:dataset_stats}
\end{figure*}

We provide summary statistics comparing the various low-quality datasets, $\mathcal{D}^L_{0.5}, \mathcal{D}^L_{0.65}, \mathcal{D}^L_{0.65}$, and our high quality dataset $\mathcal{D}^H$ in Fig.~\ref{fig:dataset_stats}. The x-axis represents the different datasets where $1$ represents $\mathcal{D}^H$. While dialogue length and Pareto-optimality are similar across all datasets, the number of unique utterances produced by Alice increases as the quality of the dataset increases. This result makes sense because low quality datasets were designed to have less unique utterances. Advantage also trends upwards with the quality of the dataset.

\section{Example Dialogues}
\label{sec:examples}
\begin{figure*}[t!]
    \centering
    \includegraphics[width=0.9\linewidth]{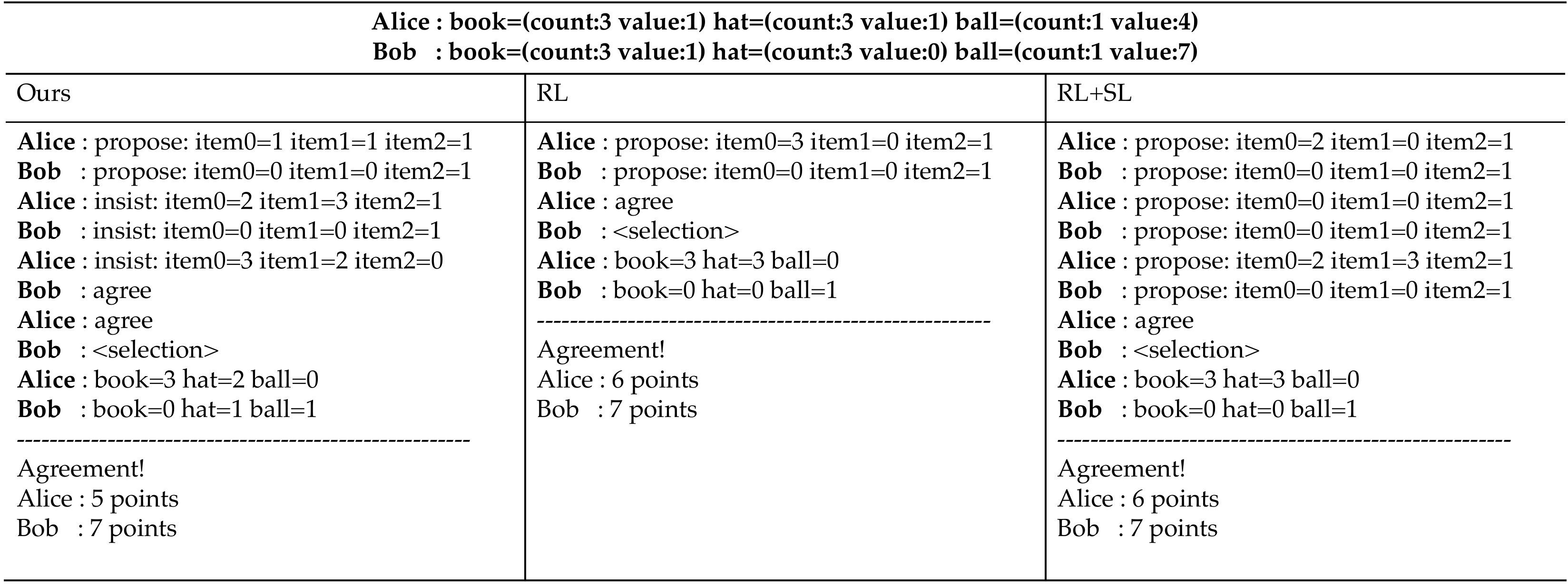}
    \vskip -0.1in
    \caption{Example dialogues where Bob is an expert agent trained on $\mathcal{D}^H$. RL Alice suggests the most aggressive proposal compared to RL+SL and our approach. } 
    \label{tab:simulated1_table}
\end{figure*}

\begin{figure*}[t!]
    \centering
    \includegraphics[width=0.9\linewidth]{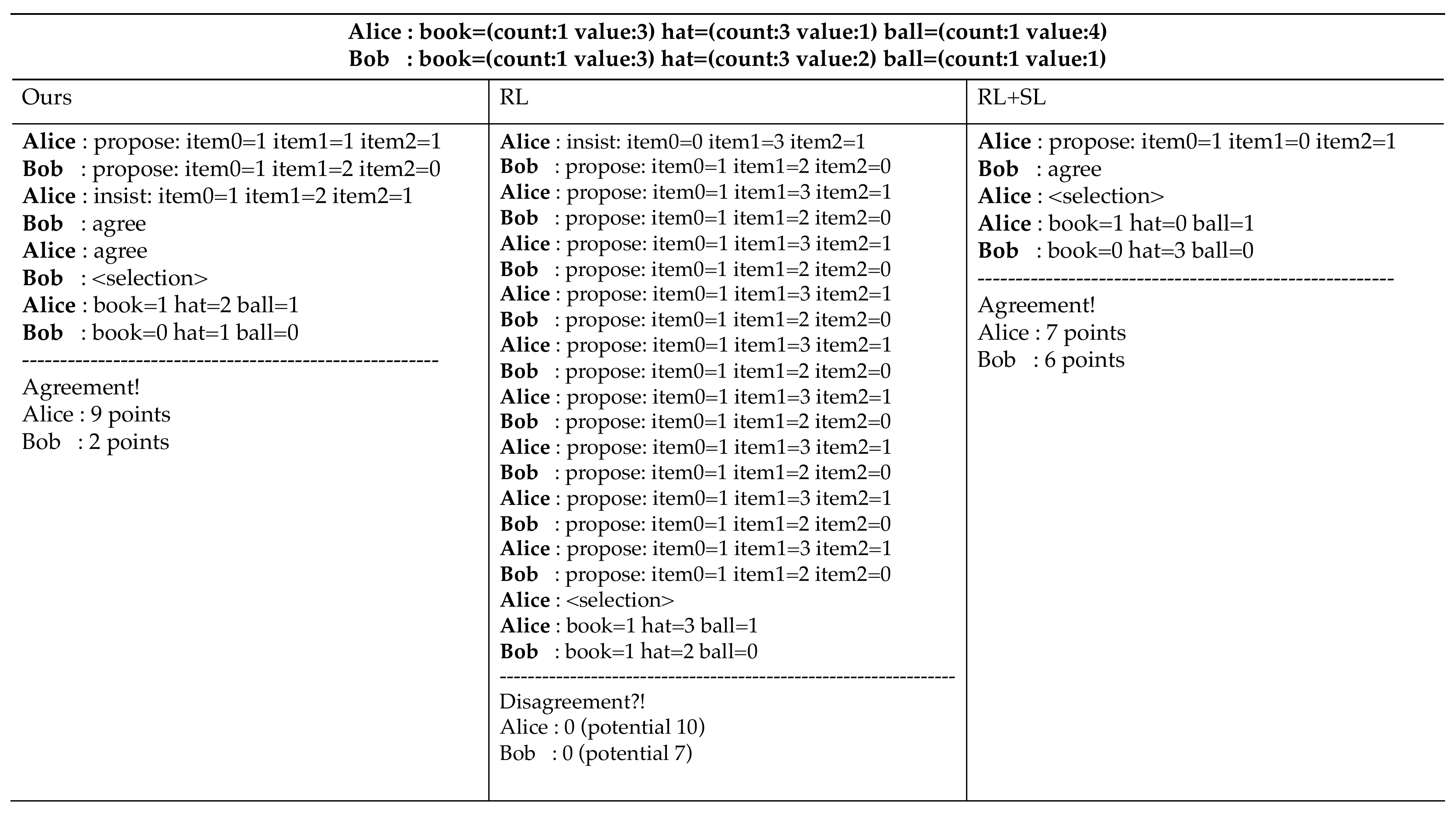}
    \vskip -0.1in
    \caption{Example dialogues where Bob is an expert agent trained on $\mathcal{D}^H$. Our Alice suggests a more advantageous proposal compared RL+SL. RL Alice displays ``badgering`` behavior where she repeatedly suggests the same (unfair) proposal. } 
    \label{tab:simulated2_table}
\end{figure*}

\begin{figure*}[t!]
    \centering
    \includegraphics[width=0.9\linewidth]{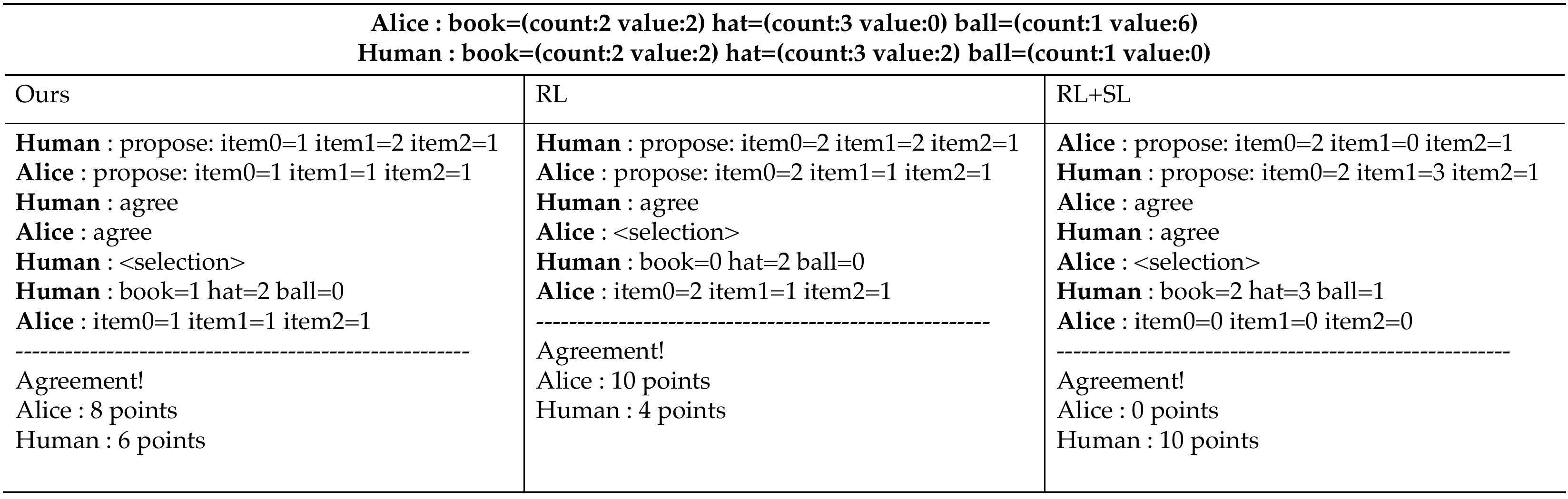}
    \vskip -0.1in
    \caption{Example dialogues against a human partner. Compared to RL and RL+SL, our Alice suggests a more compromising proposal leading to a more equitable distribution of points.} 
    \label{tab:human1_table}
\end{figure*}

\begin{figure*}[t!]
    \centering
    \includegraphics[width=0.9\linewidth]{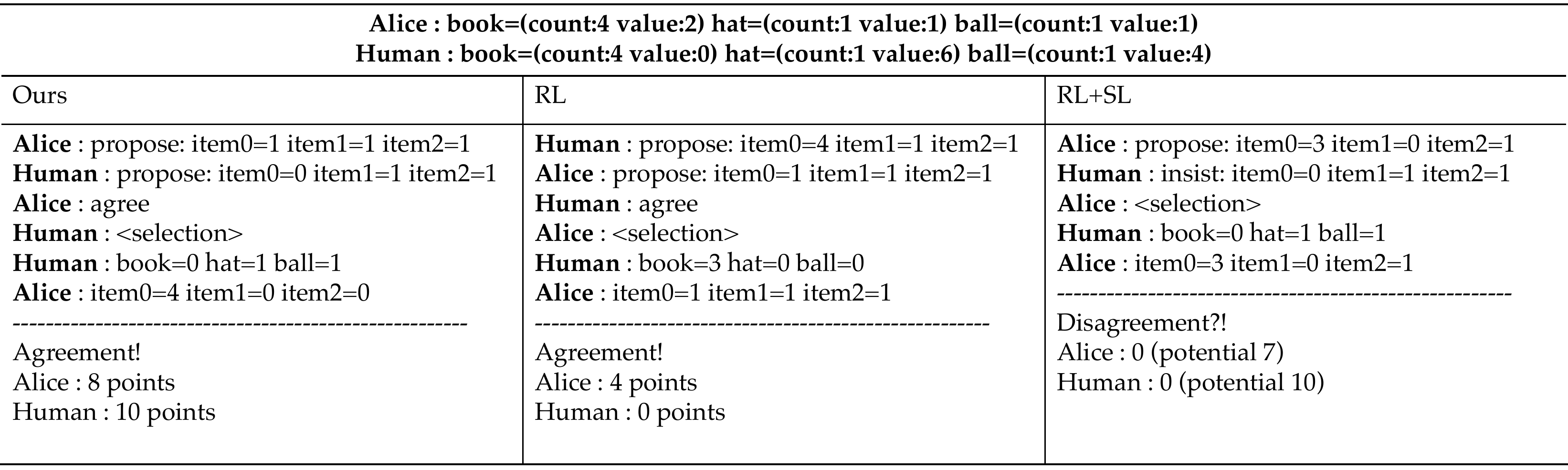}
    \vskip -0.1in
    \caption{Example dialogues against a human partner. Our Alice ends up agreeing to a more compromising proposal, resulting in scores that are highly equitable and advantageous. RL and RL+SL are not able to find equitable proposals. This results in an unfair allocation of points or a disagreement, respectively.} 
    \label{tab:human2_table}
\end{figure*}

We provide example dialogues of our models with simulated (Figs.~\ref{tab:simulated1_table}, ~\ref{tab:simulated2_table}) and human partners (Figs.~\ref{tab:human1_table}, ~\ref{tab:human2_table}). 


\end{document}